\documentclass{article}

\PassOptionsToPackage{numbers, compress}{natbib} 

\usepackage[main, final]{neurips_2026} 

\usepackage[utf8]{inputenc} 
\usepackage[T1]{fontenc}    
\ifdefined\XeTeXversion
\usepackage{newunicodechar}
\newunicodechar{’}{\textquoteright}
\newunicodechar{–}{\textendash}
\newunicodechar{—}{\textemdash}
\fi
\usepackage{hyperref}       
\usepackage{url}            
\usepackage{booktabs}       
\usepackage{amsfonts}       
\usepackage{nicefrac}       
\usepackage{microtype}      
\usepackage{xcolor}         
\usepackage{listings}
\usepackage{amsmath} 
\usepackage{multirow} 
\usepackage{enumitem} 
\usepackage{graphicx}
\usepackage{tikz}
\usepackage[most]{tcolorbox}
\usepackage{fontawesome5}
\usepackage{xcolor}
\usepackage{booktabs}
\usepackage{siunitx}

\usetikzlibrary{positioning, arrows.meta, calc}

\title{\methodname{}: Reconstruction-Guided  Reasoning Synthesis for User Modeling}

\author{
\begin{tabular}{c}
Alan Zhu\thanks{Equal contribution. Correspondence to \texttt{\{aczhu,miroyan.mihran\}@berkeley.edu}.}
\qquad
Mihran Miroyan\footnotemark[1]
\qquad
Carolyn Wang
\qquad
Andrew Zhou \\
Lisa Dunlap
\qquad
Narges Norouzi
\qquad
Joseph E. Gonzalez
\end{tabular}
\\
University of California, Berkeley
}

\newcommand{\methodname}{Recon}
\newcommand{\methodprompt}{Recon-Select}
\newcommand{\methodtrain}{Recon-GRPO}
\newcommand{\methodbaseline}{Backward Synthesis}
\newcommand{\methodbaselinetrained}{E2E-GRPO}

\definecolor{revisionpurple}{HTML}{7A1FA2}
\newif\ifshowrevisions
\showrevisionstrue

\begin{document}

\maketitle

\begin{abstract}
User modeling aims to use language models (LMs) to mimic an individual’s behavior from a corpus of past context–action pairs (e.g., conversation turns), enabling the simulation of users in settings like behavioral science, human–AI collaboration, and market research.
Recent approaches augment these corpora with synthesized reasoning traces, typically generated by conditioning on both context and action.
However, action-conditioned traces are post-hoc rationalizations and need not reflect the user's underlying decision process.
We propose \methodname{}, which uses \textit{action reconstruction} to score reasoning traces by their predictive power: given a context and candidate reasoning, a reconstruction model predicts the action, and reconstruction fidelity determines reasoning quality.
Across four domains, \methodname{} achieves a 54.7\% win rate over \methodbaseline{}, a standard post-hoc rationalization baseline. Further, we find that training a reasoning synthesis model with rewards derived from \methodname{} improves downstream user modeling performance, achieving a win rate of up to 70.0\% over baselines.
We further show that \methodname{}-synthesized reasoning transfers across models, and improves user modeling beyond the reconstruction model. Our work demonstrates that post-hoc rationalization is insufficient for reasoning synthesis, and that useful and interpretable reasoning should naturally elicit the action from the context.

\par\medskip
\begingroup
\hypersetup{hidelinks}
\centering
\small\bfseries\color[HTML]{7B5E49}
\href{https://recon-user-modeling.com}{\faWindowMaximize\enspace Project Page}
\qquad
\href{https://github.com/mmiroyan/recon-user-modeling}{\faGithub\enspace Code}
\par
\endgroup
\end{abstract}

\section{Introduction}
\label{sec:intro}

\begin{figure}[t]
    \centering
    \includegraphics[width=\columnwidth]{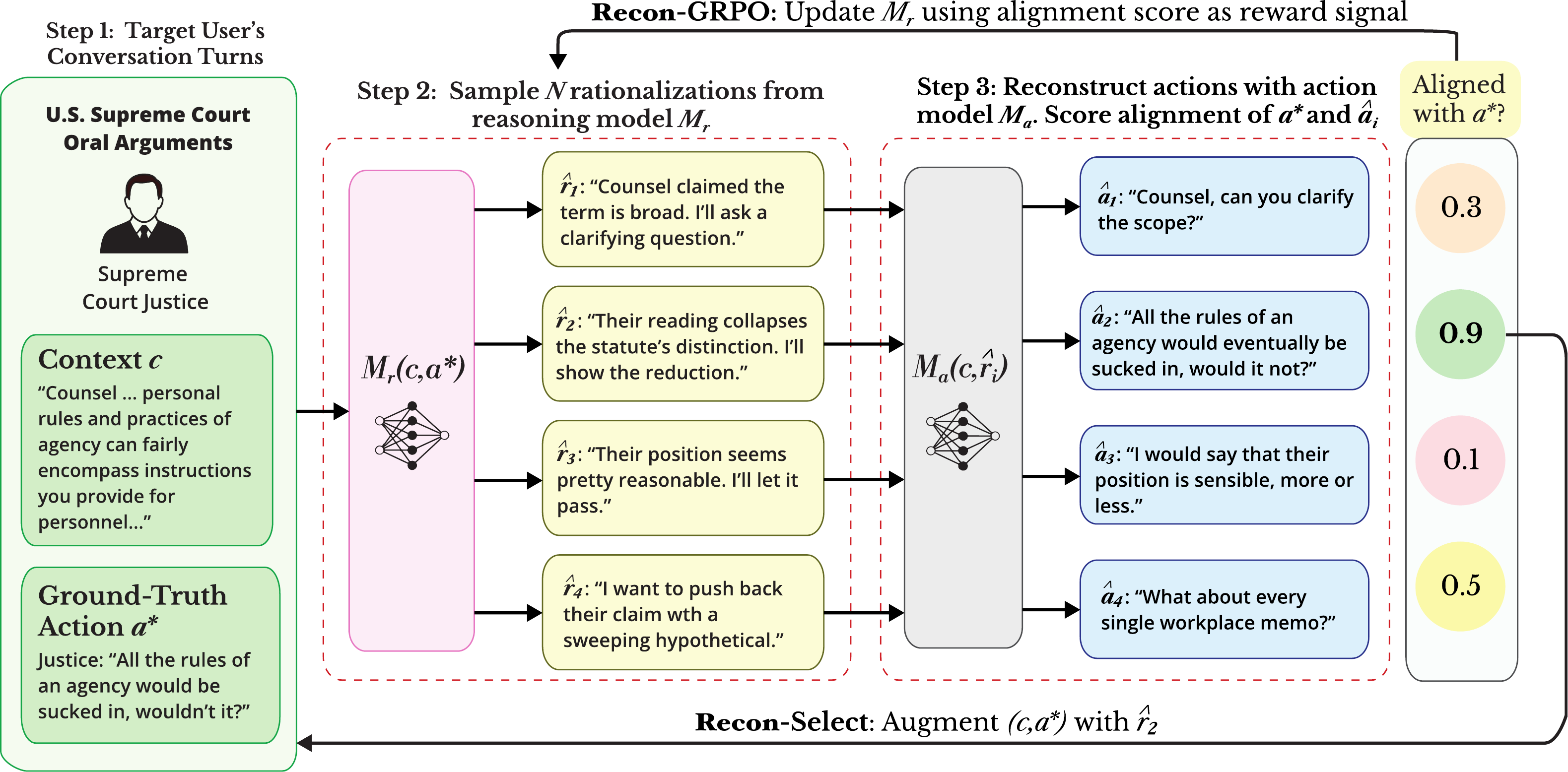}
    \caption{\textbf{\methodname{} pipeline.} Given a context--action pair $(c, a^*)$ (1), we sample $N=4$ candidate rationalizations from the reasoning model $M_r$ (2). Each candidate is evaluated by using the action model $M_a$ to reconstruct the action from the context $c$ and candidate reasoning $\hat{r}_i$, and measuring its alignment with the ground-truth action $a^*$ (3). The resulting scores are used either to select rationalizations for \methodprompt{} synthesis or as reward signals for RL training in \methodtrain{}.}
    \label{fig:main_figure}
\end{figure}

A language model (LM) that understands how a user \textit{thinks} is better able to predict how a user \textit{acts}.
LMs are increasingly used to simulate individual users in applications such as customer feedback~\citep{lu2025can}, human subjects research~\citep{hwang2025human}, and human-AI collaboration~\citep{wu2025collabllm}.
In these settings, the goal is to predict a user’s action in a given context from past interactions, which may include preference labels~\citep{ryan2025synthesizeme,singh2025fspo}, interview responses \citep{park2024generative}, or unstructured natural language interactions \citep{buening2026aligning,lu2025can,naous2025flipping}. For instance, the context may consist of prior conversation turns, and the action may be the user's next-turn response.

Recent work synthesizes reasoning traces that explain the reasoning behind an action and provide information beyond the original context–action pair, improving downstream action prediction generalization \citep{joshi-etal-2025-improving, lu2025can, naous2025flipping}.
These reasoning synthesis methods draw on approaches used in verifiable domains such as math, multi-hop QA, and coding~\citep{fang2024trace,shao2023synthetic,zhang2026steal}, where models are prompted to generate reasoning linking the context to the action.
However, these synthesized traces are post-hoc \textit{rationalizations} prone to logical leaps and arbitrary decisions.
In logic-heavy tasks, such rationalizations still produce a logically sound chain from context to answer.
However, in open-ended free-response settings like conversations, the rationalization does not necessarily lead naturally to the observed action.
As a simple example, suppose the context involves someone at a cafe, and the individual's action is ordering a glass of orange juice.
A possible rationalization is, ``I was feeling thirsty.''
Although consistent with the action, this rationale does not explain the specific choice of orange juice, since thirst would more naturally support ordering water.
A better rationale is, ``I like orange juice,'' which more directly explains the user's observed action.
Indeed, prior work in cognitive science shows that human rationalizations often diverge from underlying decision processes~\citep{nisbett1977telling}.

An alternative to post-hoc rationalization generation is to condition only on the context and retain candidate reasoning traces that produce the correct outcome, i.e., rejection sampling.
This approach can be effective in settings with small output spaces where it can be expected that with enough samples, the observed action will be covered, such as binary preferences~\citep{ryan2025synthesizeme}.
However, in open-ended natural language settings, where actions are high-dimensional, simple rejection sampling becomes infeasible. Further, prior work has explored reasoning training methods \citep{wu2026humanlm}, where a model generates both reasoning and action and is rewarded based on alignment of the generated action with the ground-truth. However, the reward depends only on the action, so the resulting reasoning traces need not be interpretable or useful for understanding user behavior, as shown in Section \ref{sec:results}.

This leaves a gap: how can we synthesize reasoning traces that are both aligned with observed actions and useful for downstream prediction?
To address this, we propose \methodname{}, which evaluates candidate rationalizations based on \emph{action reconstruction}.
We first synthesize a candidate reasoning trace conditioned on both context and ground truth action. Then, given the same context and the synthesized reasoning trace, we prompt an LM to generate a reconstructed action and measure its alignment with the ground truth.
Moreover, the rankings induced by \methodname{} over a set of candidates can be used to obtain a reward signal for training a reasoning synthesizer (\methodtrain{}). The end-to-end \methodname{} pipeline is shown in Figure \ref{fig:main_figure}.
By generating and ranking multiple candidates, \methodname{} selects rationalizations that predict user behavior, generating predictively useful, interpretable reasoning traces for in-context learning methods (\methodprompt{}). We note that \methodname{} selects among action-conditioned rationalizations; reconstruction utility does not establish that a selected trace recovers the user's true latent reasoning.

We evaluate \methodname{} across four domains: US Supreme Court oral arguments, UK Prime Minister's Questions, podcasts, and Reddit conversations. For all tasks, context consists of past turns in the conversation, and action is a spoken or written response. We compare against the reasoning synthesis method used in \citet{lu2025can} and \citet{zhang2026steal}. Methods are evaluated by synthesizing reasoning traces to augment a retrieval corpus used by a fixed retrieval-augmented generation (RAG) pipeline to generate actions for test contexts. An LM-judge performs pairwise comparison of actions generated from corpora augmented by different synthesis methods for alignment with the ground truth action; we then report win rates after excluding ties. Across four domains, \methodprompt{} improves over existing baselines with overall win rates of 54.7\% and 53.5\% using Qwen3-8B and GPT-5-mini, respectively, while \methodtrain{} further improves win rates up to 70\% with a smaller Qwen3-4B model.

Our contributions are as follows:
\begin{enumerate}
    \item We introduce \methodname{}, a method that uses action reconstruction as a criterion for evaluating synthesized reasoning in user modeling settings.
    \item We use \methodname{} to synthesize reasoning via both training-free and training-based methods, achieving win rates over existing methods of up to 70\% with training and consistently above 50\% with prompt-based synthesis.
    \item We demonstrate that selected traces can improve models not used in reconstruction, with asymmetric cross-model transfer gains.
\end{enumerate}
\section{Related Work}
\label{sec:related_work}

\subsection{User Modeling and Personalization}
\label{sec:related_personalization}

User modeling, the task of modeling an individual user’s behavior from past interactions, is closely related to personalization, which broadly aims to align models with user preferences.
Much of LM-based personalization assumes access to pairwise preference labels $(c, o^{(1)}, o^{(2)}, y)$, where $y$ indicates which output $o^{(i)}$ a user prefers in context $c$.
Personalized-RLHF~\citep{li2024personalized}, Variational Preference Learning~\citep{poddar2024personalizing}, and the Pluralistic Alignment Framework~\citep{chen2024pal} learn user-specific representations from preference labels to better align outputs to user preferences at inference time.
Personalized Soups~\citep{jang2023personalized} trains policies along pre-defined preference axes and merges their parameters, while Group Preference Optimization~\citep{zhao2023group} and Few-shot Preference Optimization~\citep{singh2025fspo} enable models to leverage preference labels in context.
In contrast, we assume only access to unstructured interaction data, which is more plentiful than controlled preference labels, to model natural-language actions directly.

Recent work has begun to study the general setting of modeling user behavior \citep{aher2023using,hwang2025human,miroyan2025parastudent,park2023generative}.
\citet{buening2026aligning} distill models from hindsight generations that incorporate follow-up user feedback, whereas we focus on modeling observed user actions directly.
HumanLM~\citep{wu2026humanlm} trains a user model with Reinforcement Learning (RL) rewards based on action alignment, but does not study reasoning synthesis as a corpus augmentation method.
Most closely related are \citet{naous2025flipping} and \citet{lu2025can}, which augment user behavior data with synthesized reasoning for supervised fine-tuning (SFT). However, both generate reasoning traces via direct post-hoc rationalization conditioned on the ground-truth action. In contrast, we use \methodname{} to evaluate and select reasoning traces based on action reconstruction fidelity, explicitly optimizing for reasoning that better guides downstream generation.

\subsection{Reasoning Synthesis}

Reasoning synthesis has been widely used to augment data in verifiable domains such as math, coding, and question answering.
STaR~\citep{zelikman2022star} tries to synthesize rationales from context alone and provides the ground-truth answer if initial rationale generation fails.
TRACE~\citep{fang2024trace} augments QA corpora by synthesizing reasoning traces over knowledge graphs.
Trace Inversion~\citep{zhang2026steal} synthesizes traces conditioned on both context and action to recover hidden reasoning behavior from closed-source models.
These methods operate in domains where correctness is externally verifiable, and reasoning follows well-defined logical steps.
In user modeling, however, reasoning traces are not uniquely determined by the observed action. As a result, post-hoc rationalizations can be consistent with an action while failing to explain why the user produced that action rather than a plausible alternative.

The closest reasoning-synthesis work to ours is SynthesizeMe~\citep{ryan2025synthesizeme}, which studies user modeling in a pairwise preference setting.
SynthesizeMe samples natural-language reasoning from context alone, uses it to predict a user's preference choice, and retains traces that recover the ground-truth choice. Joshi et al. \cite{joshi-etal-2025-improving} improve persona modeling by augmenting rationalization with psychological scaffolds, though they study structured action spaces (e.g., survey responses, preferences) instead of natural language.
We instead study unstructured natural-language interactions, where actions cannot be reduced to discrete preference choices.
Accordingly, we propose an action-reconstruction method for evaluating the quality of reasoning traces in an open-ended action space.

\section{Methodology}
\label{sec:method}

\subsection{Problem Setup}
\label{sec:problem_setup}

We define user modeling as the task of aligning a model's actions with those of a specific user in a given context $c$. We model the user as a latent policy $M_u^*$ that produces an action $a^*$ through an unobserved latent reasoning process $r^*$: $M_u^*(c, r^*(c)) \to a^*$.

A standard approach to user modeling with LMs is in-context learning, where a user's past actions are provided as in-context examples to guide action prediction in unseen contexts. These examples consist of past actions taken by the user in similar contexts. The retrieved context-action pairs can be augmented with synthesized reasoning traces to expose latent user characteristics (e.g., preferences, beliefs) and potentially improve generalization.

Concretely, given a test context $c_T$, we retrieve a set of context-action pairs $\{(c_1, a^*_1), \dots, (c_k, a^*_k)\}$ based on context similarity. For each retrieved example, a reasoning synthesis method $f$ generates a trace $\hat{r}_i = f(c_i, a^*_i)$. An action model $M_a$ is then conditioned on the augmented $(c_i, \hat{r}_i, a^*_i)$ examples and the test context, $c_T$, to generate an action:
\begin{equation}
M_a(\{(c_1, \hat{r}_1, a^*_1), \dots, (c_k, \hat{r}_k, a^*_k)\}, c_T) \to \hat{a}_T.
\label{eq:pipeline}
\end{equation}

We evaluate a method $f$ by measuring how much its synthesized traces improve the action model's ability to predict the user's ground-truth action $a_T^*$. A motivating hypothesis, explored in a model-generated setting in Appendix \ref{app:assumption_validation}, is that traces closer to the true latent reasoning improve in-context action prediction:
\begin{equation}
f(c_i, a_i^*) \approx r_i^* \Rightarrow M_a(\cdot) \approx a_T^*.
\label{eq:validity}
\end{equation}

Our goal is to improve the predictive utility of synthesized rationalizations.

\subsection{Rationalization vs. Reasoning}
\label{sec:rationalization}

We distinguish \emph{reasoning} from \emph{rationalization}: reasoning is the latent process that prompted the user to perform the observed action, i.e., elicited from $c$ and leading to $a^*$; rationalization is a post-hoc explanation obtained conditioned on the action, i.e., obtained from $(c,a^*)$.
Rationalizations are problematic as there exist many distinct reasoning traces consistent with $(c,a^*)$, so a generated trace may justify the action without reflecting the process that produced it. Recall our orange juice example from Section~\ref{sec:intro}. Both ``I was feeling thirsty,'' and ``I like orange juice'' are consistent with the action, but the latter better explains the specific choice and precludes alternatives such as ordering water. This distinction is important for user modeling: good reasoning reveals latent user characteristics (e.g., a preference for orange juice) and focuses options, supporting more accurate downstream predictions. However, existing reasoning synthesis methods typically rely on rationalization.

An alternative to rationalization is to generate reasoning-and-action traces from $c$ until an action aligned with $a^*$ is generated: the more accurate the action, the better the reasoning.
In our setting, however, rejection sampling is infeasible as the action space consists of free-form natural language.

\subsection{\methodname{}}
\label{sec:reconstruction}

Inspired by Section \ref{sec:rationalization}'s insights, we use action reconstruction, or \textbf{\methodname{}}, as a tractable surrogate for the rejection-sampling method above. Instead of sampling reasoning-and-action traces from $c$ until the generated action matches $a^*$, \methodname{} starts from rationalizations generated from $(c,a^*)$ and tests whether each rationalization can still recover the action when the action itself is withheld. Given a candidate rationalization, $\hat{r}$, we condition the action model, $M_a$, on only the context and rationalization: $\hat{a} = M_a(c, \hat{r})$. The quality of $\hat{r}$ is quantified by the alignment between the reconstructed action $\hat{a}$ and the observed action $a^*$. To prevent reconstruction from rewarding trivial copying, the synthesis model is instructed not to explicitly reveal the observed action in its generation.

We use \methodname{} in two ways. First, we propose \textbf{\methodprompt{}}, a training-free approach that uses reconstruction to select among candidate rationalizations. Let $f$ denote a reasoning synthesis method, i.e., $f(c,a^*) \rightarrow \hat{r}$. To synthesize better reasoning, we sample $N$ candidate traces $\{\hat{r}_i\}_{i=1}^{N} \sim f(c,a^*)$, reconstruct the corresponding actions $\hat{a}_i = M_a(c,\hat{r}_i)$, and select the trace $\hat{r}_j$ whose reconstructed action $\hat{a}_j$ is most aligned with $a^*$. We follow HumanLM~\citep{wu2026humanlm} and use an LM judge to score alignment between reconstructed and ground-truth actions along three predefined dimensions (style, intent, and values) before making an overall evaluation.

Second, we explore whether the same reconstruction score can serve as a reward signal for RL training, \textbf{\methodtrain{}}. Specifically, for a given rollout $\hat{r} = M_r(c, a^*)$, the reward is defined as the alignment of the reconstructed action $\hat{a}=M_a(c,\hat{r})$ with $a^*$. This training method is closely related to \textsc{Advisor Models}~\citep{asawa2025train}, which trains an auxiliary model to prompt (i.e., advise) a frozen model to improve its performance on a downstream task. In our setting, $M_r$ serves as an advisor to the action model $M_a$: it learns to produce reasoning traces that facilitate the reconstruction of the action. By keeping $M_a$ frozen, we ``force'' the reasoning model to output interpretable and meaningful reasoning traces, rather than learning to generate accurate actions.

We use Gemini-3.1-Flash-Lite~\citep{google2026gemini31flashlite} as the judge model in both \methodprompt{} and \methodtrain{}; both pipelines are shown in Figure \ref{fig:main_figure}. Further details are in Appendix \ref{app:method_details}.

\section{Experimental Setup}
\label{sec:experiments}

\subsection{Data}

\begin{figure}[t]
    \centering
    \includegraphics[width=0.95\columnwidth]{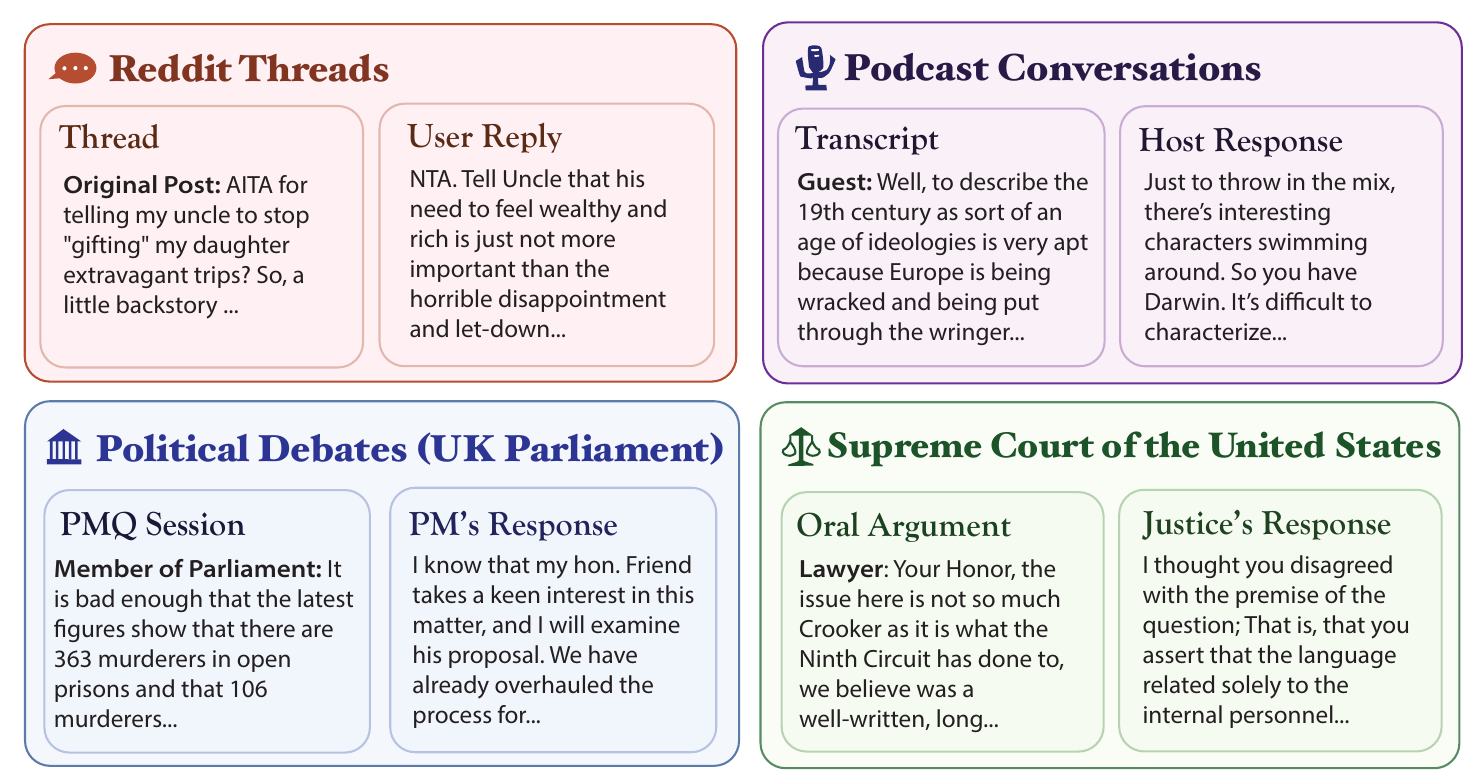}
    \caption{\textbf{Data domains.} We study Reddit, Podcasts, U.K. Parliament debates, and U.S. Supreme Court oral arguments, ranging from brief spoken formal questions to long-form informal writing.}
    \label{fig:data_figure}
    \vspace{-0.15in}
\end{figure}

Consistent with our setup in Section \ref{sec:method}, we focus on domains where the action space is free-form natural language, specifically spoken and written conversations. Context consists of prior turns of the conversation, and the observed action is an individual's next-turn utterance. The downstream task is to predict the individual's utterance given a test context. We consider four domains:

\textbf{Supreme Court of the United States (SCOTUS).} We model two U.S. Supreme Court Justices during oral arguments, where lawyers present cases before the Court and Justices probe legal arguments.

\textbf{Prime Minister's Questions (PMQ).} We model two U.K. Prime Ministers (PMs) during PMQ sessions, where Parliament questions the PM on current events and government policies and decisions.

\textbf{Podcasts.} We model two podcast hosts across long-form interview episodes, where two-person discussions cover a wide range of topics and include both questions and opinionated commentary.

\textbf{Reddit.} We model two Reddit users from an online community where commenters evaluate interpersonal disputes by assigning judgments and explaining their reasoning.

Example context--action samples for the above domains are provided in Figure \ref{fig:data_figure}. We hold out test conversations, with a test split of 512 context--action pairs. Additional data statistics and details on the data processing pipeline are in Appendix~\ref{app:data}.

\subsection{Baselines}

Following prior work \citep{lu2025can, zhang2026steal}, we define a baseline reasoning synthesis method \textbf{\methodbaseline{}}, where for each context-action pair $(c_i, a_i^*)$ in the corpus we synthesize a single reasoning trace by prompting a reasoning model $M_r$ conditioned on $(c_i, a_i^*)$.

We also define a trained baseline \textbf{\methodbaselinetrained{}} (end-to-end GRPO). Following prior training work~\citep{wu2026humanlm}, we train $M_r$ with Group Relative Policy Optimization (GRPO)~\citep{shao2024deepseekmath} to generate both reasoning and action from the context, with rewards based solely on the generated action's alignment to the ground-truth action scored by an LM judge.

\subsection{Experiments}
\label{subsec:evaluation}

We evaluate \methodprompt{} by sampling $N=4$ candidates per retrieved context--action pair from \methodbaseline{}. Unless otherwise stated, the same model is used for reasoning synthesis $M_r$ and action generation $M_a$. We run this setup with Qwen3-8B~\citep{yang2025qwen3} and GPT-5-mini~\citep{singh2025openai}.
For the trained baseline \methodbaselinetrained{} and \methodtrain{}, we use Low Rank Adaptation (LoRA)~\citep{hu2022lora} for GRPO fine-tuning with rank $128$ and group size $4$; because GPT-5-mini is not trainable in our setup, we evaluate train-based methods only with Qwen3-8B. Implementation details are provided in Appendix~\ref{app:method_details}.

Performance is measured as win rate against \methodbaseline{} in the retrieval-augmented action-generation pipeline from Equation~\ref{eq:pipeline}. For each test example, the compared methods use the same retrieved examples, test context, and action model; only the reasoning augmentations differ. We follow HumanLM~\citep{wu2026humanlm} and use Gemini-3.1-Flash-Lite to compare each generated action to the ground-truth action along three predefined dimensions (style, intent, and values) before making an overall evaluation. We report per-dimension and overall win rates after excluding ties. Additional evaluation details are provided in Appendix~\ref{app:evaluation_details}. To validate the automatic evaluation, two blind human raters evaluate a set of 100 response pairs, giving \methodprompt{} a 60.7\% mean non-tied win rate (Appendix~\ref{subsec:human_validation}). We additionally evaluate using an LM from a different model family (Appendix~\ref{subsec:disjoint_validation}).

\section{Results}
\label{sec:results}

\begin{figure}[tbp]
    \centering
    \includegraphics[width=\columnwidth]{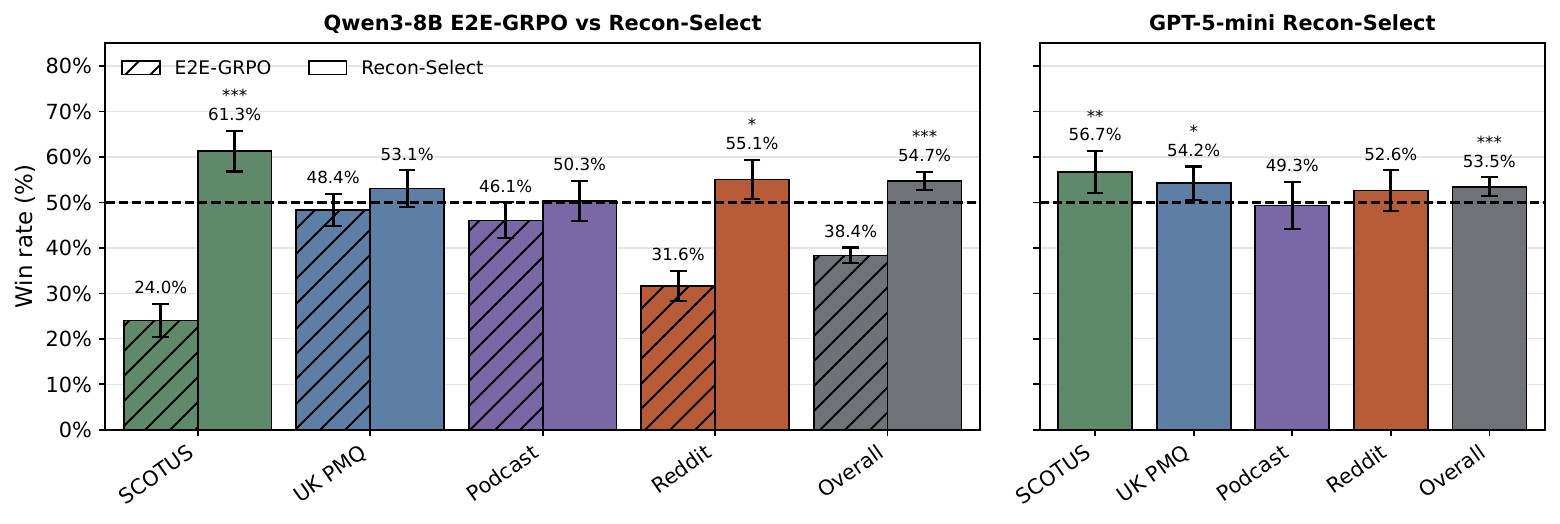}
    \caption{
        \textbf{\methodprompt{} and \methodbaselinetrained{} results.}
        Win rates against \methodbaseline{} across four domains and overall.
        \textbf{(Left)} On Qwen3-8B, \methodprompt{} achieves a 54.7\% overall win rate over baseline. \methodbaselinetrained{} achieves only a 38.4\% win rate, indicating that optimizing for action accuracy alone does not produce transferable reasoning traces.
        \textbf{(Right)} On GPT-5-mini, \methodprompt{} achieves a 53.5\% overall win rate, demonstrating consistent improvements from reconstruction-based selection.
        Asterisks denote win rates significantly exceeding 50\% ($^*p<0.05$, $^{**}p<0.01$, $^{***}p<0.001$). 95\% confidence intervals shown.
        }
    \label{fig:recon}
\end{figure}
\begin{figure}[htbp]
    \centering
    \includegraphics[width=\columnwidth]{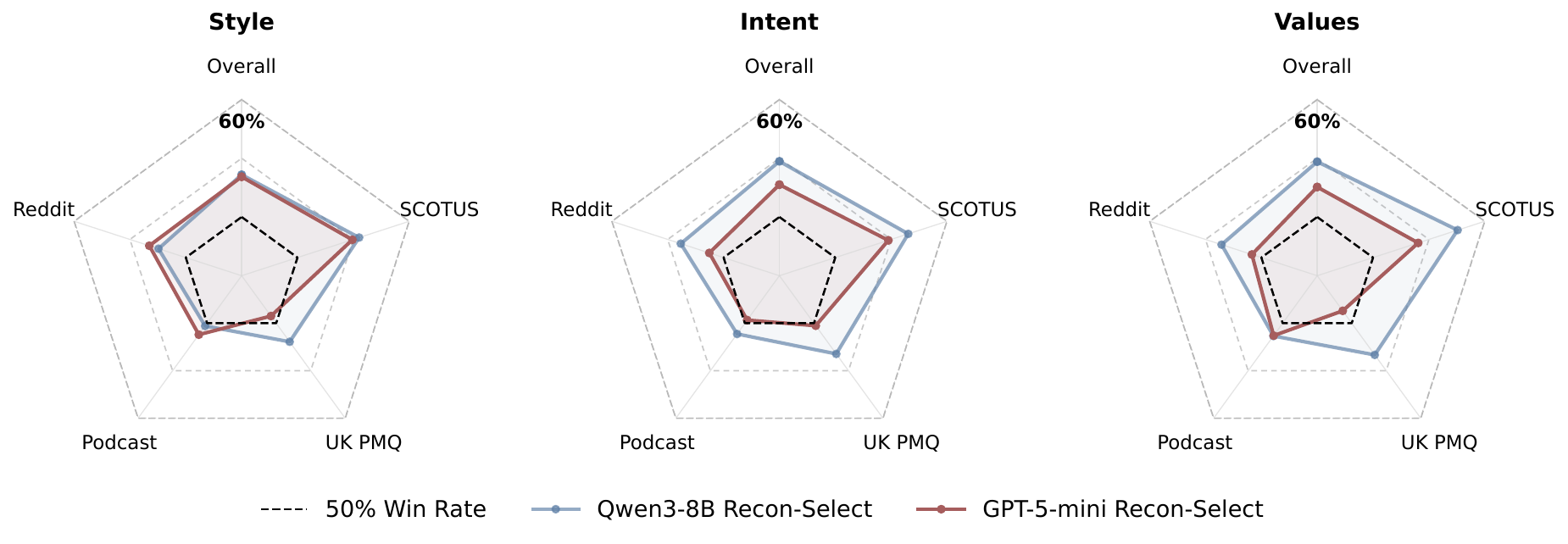}
    \caption{\textbf{\methodprompt{} results by dimension.} Win rates of \methodprompt{} against \methodbaseline{} across domains and overall broken down by alignment dimensions: Style \textbf{(Left)}, Intent \textbf{(Middle)}, and Values \textbf{(Right)}. \methodprompt{} consistently improves over the baseline for both Qwen3-8B and GPT-5-mini models across all dimensions. The inner polygon denotes 50\% win rate.
    }
    \label{fig:dimensions}
\end{figure}

\methodprompt{}’s win rates against \methodbaseline{} are shown on the left (Qwen3-8B) and right (GPT-5-mini) of Figure~\ref{fig:recon}. \methodprompt{} reliably outperforms Backward Synthesis, achieving overall win rates of 54.7\% in the Qwen3-8B setting and 53.5\% in the GPT-5-mini setting. Notably, no domain exhibits a win rate significantly below 50\%, providing strong evidence that \methodname{}-based selection reliably improves over existing reasoning synthesis methods. While the absolute gains may appear modest, they are notable given the difficulty of user modeling in noisy, high-dimensional natural language action spaces, and are consistent with prior work demonstrating the inherent challenge of predicting user behavior from unstructured data \citep{wu2026humanlm}.

We validate the results using an LM judge from another model family. Specifically, using Gemini-3.1-Flash-Lite only for trace selection, GPT-5.6 Terra as a judge gives Qwen3-8B \methodprompt{} a 52.1\% pooled win rate (95\% bootstrapped CI: 50.4-53.9\%). Additional details and validation ablations are provided in Appendix~\ref{subsec:human_validation}--\ref{subsec:rebuttal_controls}.

We further break down \methodprompt{}'s improvements by the judge's alignment dimensions in Figure~\ref{fig:dimensions}. Across style, intent, and values, \methodprompt{} generally maintains win rates above 50\% for both Qwen3-8B and GPT-5-mini, indicating that gains are not driven by a single evaluation criterion. This suggests that reconstruction-based selection helps recover higher-level user characteristics across latent dimensions, including not only surface style but also communicative intent and values.

The left side of Figure~\ref{fig:recon} shows \methodbaselinetrained{}’s performance against \methodbaseline{}. Surprisingly, training the reasoning synthesis model directly for action accuracy substantially \emph{degrades} downstream performance, achieving only a 38.4\% win rate overall and significantly underperforming in three of four domains. These results provide strong evidence that reasoning optimized solely to reproduce actions does not necessarily yield transferable reasoning traces. Instead, useful reasoning must be selected based on demonstrated downstream utility rather than model-specific action accuracy alone. In contrast, Section \ref{subsec:transferability} shows that reasoning traces selected by \methodprompt{} successfully \emph{transfer} across models, improving user modeling performance beyond the reconstruction model itself.

To show the utility of persona-specific reasoning synthesis, we ran a cross-persona control in which retrieved examples and \methodprompt{} reasoning come from one persona, while the active context and prediction target come from another. On SCOTUS with Qwen3-8B, the win rate falls to 49.0\%, versus 61.3\% in the matched setting, indicating user-specific signal in the synthesized reasoning traces through the reconstruction objective.

\subsection{Transferability of \methodprompt{} Reasoning}
\label{subsec:transferability}

\begin{figure}[tp]
    \centering
    \includegraphics[width=0.80\columnwidth]{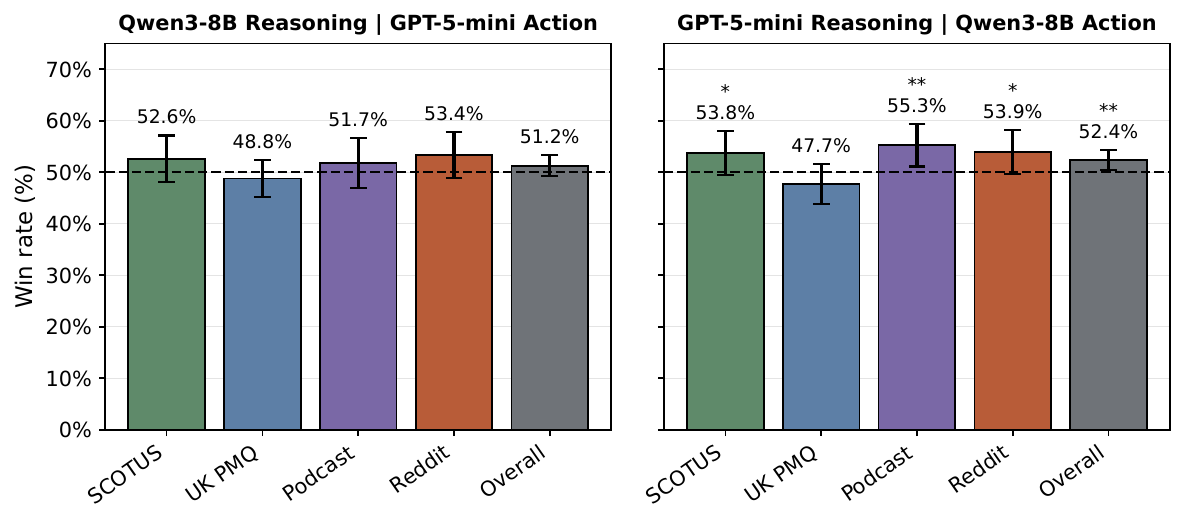}
    \caption{
        \textbf{Cross-model transfer results.}
        Win rates for \methodprompt{} against \methodbaseline{} using different reasoning ($M_r$) and action ($M_a$) models.
        \textbf{(Left)} Qwen3-8B as $M_r$ and GPT-5-mini as $M_a$: gains are not statistically significant, suggesting limited transfer from weaker to stronger models.
        \textbf{(Right)} GPT-5-mini as $M_r$ and Qwen3-8B as $M_a$: gains are statistically significant, demonstrating that \methodprompt{} reasoning traces transfer across model families.
        Asterisks denote win rates significantly exceeding 50\% ($^*p<0.05$, $^{**}p<0.01$). 95\% confidence intervals shown.
    }
    \label{fig:cross_model}
\end{figure}

To evaluate the transferability and interpretability of reasoning synthesized via \methodprompt{}, we swap the reasoning and action-generation models used in our pipelines: we provide corpora augmented using Qwen3-8B to GPT-5-mini action generation and vice versa, comparing against corpora augmented with \methodbaseline{}. These experiments test whether reasoning selected to guide one model remains useful for guiding another. Results are shown in Figure~\ref{fig:cross_model}.

When GPT-5-mini generates actions using reasoning synthesized with Qwen3-8B, \methodprompt{} does not significantly beat \methodbaseline{}, suggesting limited transfer from weaker to stronger models. This asymmetry is consistent with the stronger model already generating high-quality actions from the context alone, leaving less room for weaker reasoning traces to add useful signal. In contrast, when Qwen3-8B uses reasoning synthesized with GPT-5-mini, \methodprompt{} achieves a statistically significant 52.4\% win rate over baseline, showing GPT-5-mini produces reasoning interpretable by Qwen3-8B. Though these gains are smaller than when the reasoning and action-generation models match, the results demonstrate \methodprompt{} selects reasoning traces that transfer across models and remain useful beyond the reconstruction model itself.

\subsection{Reasoning Model Ablations}

\begin{figure}[tbp]
    \centering
    \includegraphics[width=0.90\columnwidth]{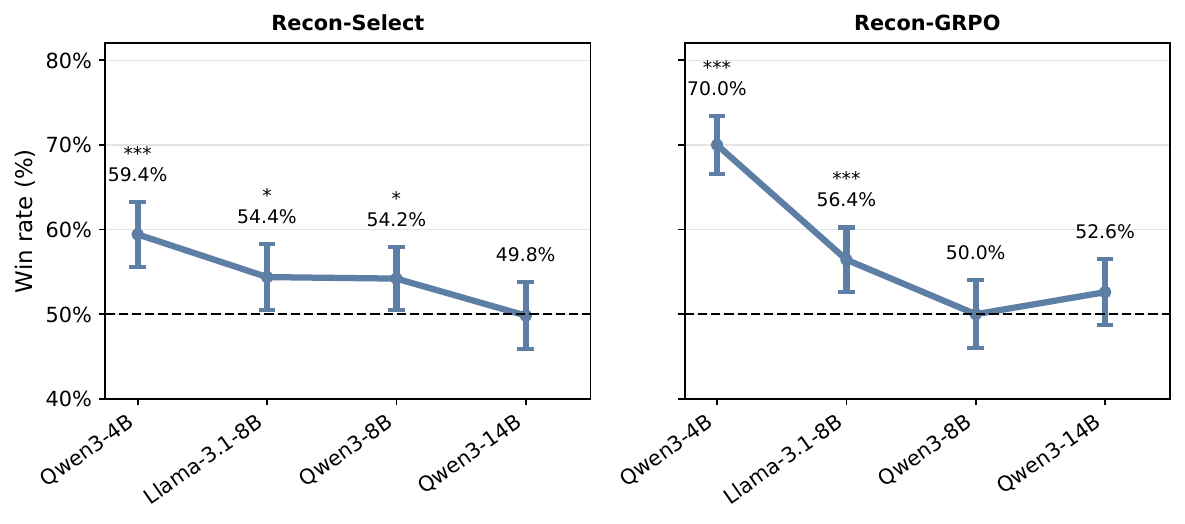}
    \caption{\textbf{\methodprompt{} and \methodtrain{} results across models.} Win rates against \methodbaseline{} on the PMQ domain across $M_r$ model families and sizes (Qwen3-4B, Llama-3.1-8B-Instruct, Qwen3-8B, Qwen3-14B), with Qwen3-8B fixed as $M_a$. \textbf{(Left)} \methodprompt{}: weaker reasoning models benefit most, while stronger models leave less headroom for improvement. \textbf{(Right)} \methodtrain{}: RL training with \methodname{}-based rewards yields further gains over \methodprompt{} on average, with the largest improvements for smaller models. Asterisks denote statistical significance of win rates exceeding 50\% ($^*p<0.05$, $^{**}p<0.01$, $^{***}p<0.001$). 95\% confidence intervals shown.}
    \label{fig:model_ablations}
    \vspace{-0.15in}
\end{figure}

To demonstrate generalization, we further evaluate \methodprompt{} against \methodbaseline{} with other reasoning generators. To control costs and enable us to test more models, we focus on the Prime Minister's Questions (PMQ) domain. We retain Qwen3-8B as the action generator and evaluate Qwen3-4B, Qwen3-8B, Qwen3-14B, and Llama-3.1-8B-Instruct as the reasoning generator.

The results are shown in Figure~\ref{fig:model_ablations}. \methodprompt{} yields larger gains for weaker reasoning models: Qwen3-4B reaches a 59.4\% win rate over \methodbaseline{}, while Llama-3.1-8B-Instruct and Qwen3-8B reach 54.4\% and 54.2\%, respectively. In contrast, Qwen3-14B shows no significant improvement. This pattern suggests that stronger models already generate higher-quality candidate reasoning traces, leaving less room for selection to improve performance, especially with only $N=4$ candidates; we provide scaling ablations in Appendix~\ref{subsec:rebuttal_controls}.

\subsection{\methodname{}-Guided Training}
\label{sec:trained_results}

\begin{figure}[htbp]
    \centering
    \includegraphics[width=0.95\columnwidth]{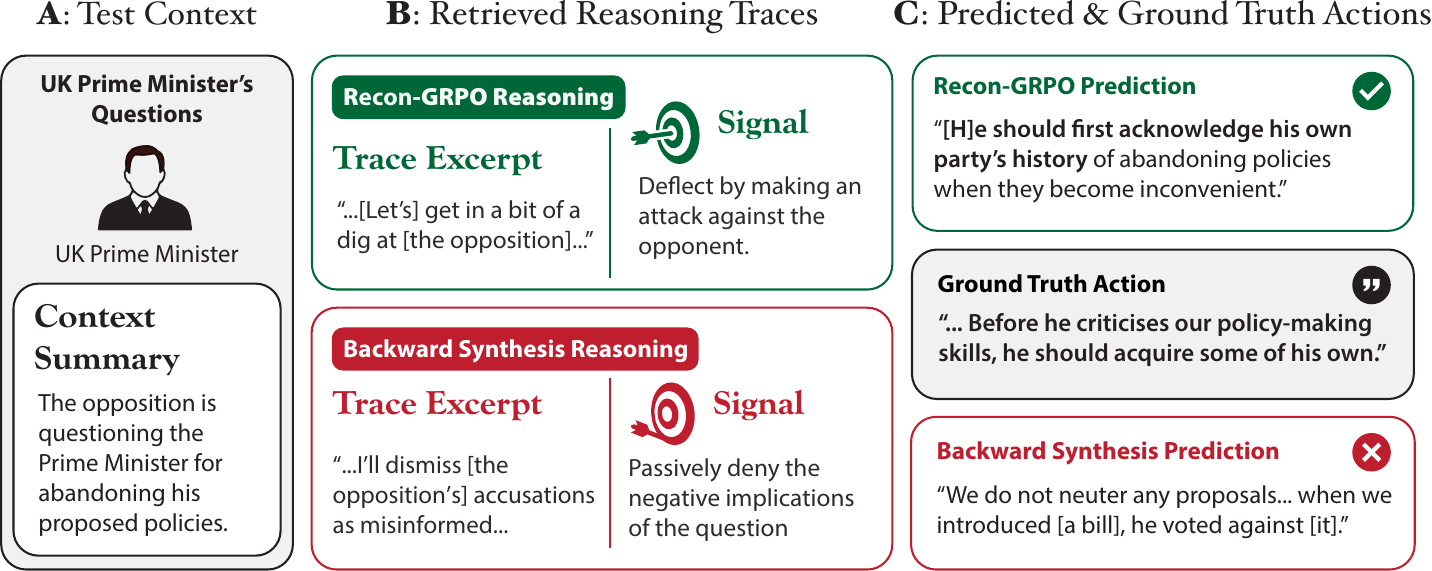}
    \caption{\textbf{Qualitative example from PMQ.} \methodtrain{} identifies the Prime Minister's intended move as attacking the opposition rather than responding defensively, guiding the action model toward a response closer to the ground truth. In contrast, Backward Synthesis produces a defensive rationale, leading to a less aligned prediction.
    }
\label{fig:qualitative}
\end{figure}

The results of \methodtrain{} against \methodbaseline{} on the model ablation experiments are on the right of Figure~\ref{fig:model_ablations}. For both Qwen3-4B and Llama-3.1-8B-Instruct, we observe higher win rates than those provided by \methodprompt{} at 70.0\% and 56.4\%, respectively, despite \methodtrain{} generating only one reasoning candidate per retrieved context-action pair (\methodprompt{} generated $N=4$ candidates). The win rate for Qwen3-14B is 52.6\%, an improvement over the \methodprompt{}'s win rate of 49.8\%, though still not large enough to make strong statistical claims at our sample size. 

The main exception is Qwen3-8B, where \methodtrain{} obtains a 50.0\% win rate. One possible explanation is that using Qwen3-8B as both the reasoning policy and the frozen reconstructor reduces reward contrast across rollouts as reasoning traces are in-domain for the action model, making the GRPO signal less informative. We leave a fuller analysis of this effect to future work.

Qualitatively, Figure~\ref{fig:qualitative} illustrates how \methodtrain{} can learn reasoning traces that better capture the user's communicative behavior. In the example from the PMQ task, \methodtrain{} identifies the Prime Minister's tendency to attack the opposition rather than respond defensively, guiding the action model toward a response closer to the ground-truth action.

\subsection{Limitations}
\label{sec:limitations}

We recognize some limitations of our work. First, we evaluate \methodname{} primarily in the user simulation setting, demonstrating effectiveness across eight individuals in four domains, and do not study its utility in verifiable domains, such as math and coding. Second, our exploration of \methodtrain{} is limited to a single domain, though the results suggest promise for \methodname{}-based learning objectives. Third, \methodname{} introduces additional computational cost: \methodprompt{} with branching factor $N$ requires $2N+1$ LM calls, and \methodtrain{} scales with group size and GPU usage, although these costs are amortized over the lifetime of the corpus. Finally, while we distinguish between rationalization and reasoning, \methodname{} still operates on rationalizations conditioned on the ground-truth action; developing methods that enable reasoning generation conditioned only on context remains an important direction for future work. Furthermore, withholding curated data and trained models limits reproducibility, and modeling identifiable individuals raises privacy and consent concerns (Appendix~\ref{app:ethics_statement}).

\section{Conclusion}
\label{sec:conclusion}

\methodname{} evaluates synthesized reasoning by testing whether it enables reconstruction of a user’s observed action from the original context. Building on this reconstruction signal, \methodprompt{} selects stronger reasoning traces without training, achieving up to a 54.7\% win rate over existing reasoning synthesis methods across domains. \methodtrain{} further uses the same signal as a reward for training reasoning synthesizers, improving win rates up to 70.0\%. More fundamentally, our results suggest that useful reasoning — the kind that predicts and explains human behavior — must be evaluated by its predictive utility, not its surface consistency with observed actions. Reconstruction fidelity is one tractable instantiation of this principle; we expect this principle to generalize to other settings where reasoning synthesis is currently dominated by post-hoc rationalization.

\begin{ack}
Sky Computing Lab is supported by gifts from Accenture, Algorithmic SuperIntelligence Labs, Amazon, AMD, Anyscale, Broadcom, cmpnd, Google, IBM, Intel, Intesa Sanpaolo, Lambda, Lightspeed, Mirendil, NVIDIA, Soma Labs, and VESSL.

This material is based upon work supported by the National Science Foundation under Grant No. DGE 2146752. Any opinions, findings, and conclusions or recommendations expressed in this material are those of the author(s) and do not necessarily reflect the views of the National Science Foundation.
\end{ack}

\newpage
\bibliography{main}

@article{lu2025can,
  title={Can {LLM} Agents Simulate Multi-Turn Human Behavior? Evidence from Real Online Customer Behavior Data},
  author={Lu, Yuxuan and Huang, Jing and Han, Yan and Yao, Bingsheng and Bei, Sisong and Gesi, Jiri and Xie, Yaochen and He, Qi and Wang, Dakuo and others},
  journal={arXiv preprint arXiv:2503.20749},
  year={2025}
}

@article{wu2026humanlm,
  title={{HumanLM}: Simulating Users with State Alignment Beats Response Imitation},
  author={Wu, Shirley and Choi, Evelyn and Khatua, Arpandeep and Wang, Zhanghan and He-Yueya, Joy and Weerasooriya, Tharindu Cyril and Wei, Wei and Yang, Diyi and Leskovec, Jure and Zou, James},
  journal={arXiv preprint arXiv:2603.03303},
  year={2026}
}

@inproceedings{ryan2025synthesizeme,
  title={{SynthesizeMe}! inducing persona-guided prompts for personalized reward models in {LLMs}},
  author={Ryan, Michael J and Shaikh, Omar and Bhagirath, Aditri and Frees, Daniel and Held, William Barr and Yang, Diyi},
  booktitle={Proceedings of the 63rd Annual Meeting of the Association for Computational Linguistics (Volume 1: Long Papers)},
  pages={8045--8078},
  year={2025}
}

@article{poddar2024personalizing,
  title={Personalizing reinforcement learning from human feedback with variational preference learning},
  author={Poddar, Sriyash and Wan, Yanming and Ivison, Hamish and Gupta, Abhishek and Jaques, Natasha},
  journal={Advances in Neural Information Processing Systems},
  volume={37},
  pages={52516--52544},
  year={2024}
}

@article{li2024personalized,
  title={Personalized language modeling from personalized human feedback},
  author={Li, Xinyu and Zhou, Ruiyang and Lipton, Zachary C and Leqi, Liu},
  journal={arXiv preprint arXiv:2402.05133},
  year={2024}
}

@article{jang2023personalized,
  title={Personalized soups: Personalized large language model alignment via post-hoc parameter merging},
  author={Jang, Joel and Kim, Seungone and Lin, Bill Yuchen and Wang, Yizhong and Hessel, Jack and Zettlemoyer, Luke and Hajishirzi, Hannaneh and Choi, Yejin and Ammanabrolu, Prithviraj},
  journal={arXiv preprint arXiv:2310.11564},
  year={2023}
}

@inproceedings{zhao2023group,
  title={Group Preference Optimization: Few-Shot Alignment of Large Language Models},
  author={Zhao, Siyan and Dang, John and Grover, Aditya},
  booktitle={NeurIPS 2023 Workshop on Instruction Tuning and Instruction Following},
  year={2023}
}

@inproceedings{hwang2025human,
  title={Human subjects research in the age of generative {AI}: Opportunities and challenges of applying {LLM}-simulated data to {HCI} studies},
  author={Hwang, Angel Hsing-Chi and Bernstein, Michael S and Sundar, S Shyam and Zhang, Renwen and Horta Ribeiro, Manoel and Lu, Yingdan and Chang, Serina and Wu, Tongshuang and Yang, Aimei and Williams, Dmitri and others},
  booktitle={Proceedings of the Extended Abstracts of the CHI Conference on Human Factors in Computing Systems},
  pages={1--7},
  year={2025}
}

@inproceedings{chen2024pal,
  title={{PAL}: Pluralistic Alignment Framework for Learning from Heterogeneous Preferences},
  author={Chen, Daiwei and Chen, Yi and Rege, Aniket and Vinayak, Ramya Korlakai},
  booktitle={NeurIPS 2024 Workshop on Behavioral Machine Learning},
  year={2024}
}

@article{wu2025collabllm,
  title={{CollabLLM}: From passive responders to active collaborators},
  author={Wu, Shirley and Galley, Michel and Peng, Baolin and Cheng, Hao and Li, Gavin and Dou, Yao and Cai, Weixin and Zou, James and Leskovec, Jure and Gao, Jianfeng},
  journal={arXiv preprint arXiv:2502.00640},
  year={2025}
}

@article{park2024generative,
  title={Generative agent simulations of 1,000 people},
  author={Park, Joon Sung and Zou, Carolyn Q and Shaw, Aaron and Hill, Benjamin Mako and Cai, Carrie and Morris, Meredith Ringel and Willer, Robb and Liang, Percy and Bernstein, Michael S},
  journal={arXiv preprint arXiv:2411.10109},
  volume={52},
  year={2024}
}

@article{naous2025flipping,
  title={Flipping the dialogue: Training and evaluating user language models},
  author={Naous, Tarek and Laban, Philippe and Xu, Wei and Neville, Jennifer},
  journal={arXiv preprint arXiv:2510.06552},
  year={2025}
}

@article{singh2025fspo,
  title={{FSPO}: Few-Shot Optimization of Synthetic Preferences Personalizes to Real Users},
  author={Singh, Anikait and Hsu, Sheryl and Hsu, Kyle and Mitchell, Eric and Ermon, Stefano and Hashimoto, Tatsunori and Sharma, Archit and Finn, Chelsea},
  journal={arXiv preprint arXiv:2502.19312},
  year={2025}
}

@article{buening2026aligning,
  title={Aligning language models from user interactions},
  author={Buening, Thomas Kleine and H{\"u}botter, Jonas and P{\'a}sztor, Barna and Shenfeld, Idan and Ramponi, Giorgia and Krause, Andreas},
  journal={arXiv preprint arXiv:2603.12273},
  year={2026}
}

@inproceedings{shao2023synthetic,
  title={Synthetic prompting: Generating chain-of-thought demonstrations for large language models},
  author={Shao, Zhihong and Gong, Yeyun and Shen, Yelong and Huang, Minlie and Duan, Nan and Chen, Weizhu},
  booktitle={International conference on machine learning},
  pages={30706--30775},
  year={2023},
  organization={PMLR}
}

@inproceedings{fang2024trace,
  title={{TRACE} the evidence: Constructing knowledge-grounded reasoning chains for retrieval-augmented generation},
  author={Fang, Jinyuan and Meng, Zaiqiao and Macdonald, Craig},
  booktitle={Findings of the Association for Computational Linguistics: EMNLP 2024},
  pages={8472--8494},
  year={2024}
}

@article{zhang2026steal,
  title={How to Steal Reasoning Without Reasoning Traces},
  author={Zhang, Tingwei and Morris, John X and Shmatikov, Vitaly},
  journal={arXiv preprint arXiv:2603.07267},
  year={2026}
}

@article{nisbett1977telling,
  title={Telling more than we can know: Verbal reports on mental processes.},
  author={Nisbett, Richard E and Wilson, Timothy D},
  journal={Psychological review},
  volume={84},
  number={3},
  pages={231},
  year={1977},
  publisher={American Psychological Association}
}

@article{shao2024deepseekmath,
  title={{DeepSeekMath}: Pushing the limits of mathematical reasoning in open language models},
  author={Shao, Zhihong and Wang, Peiyi and Zhu, Qihao and Xu, Runxin and Song, Junxiao and Bi, Xiao and Zhang, Haowei and Zhang, Mingchuan and Li, YK and others},
  journal={arXiv preprint arXiv:2402.03300},
  year={2024}
}

@article{asawa2025train,
  title={How to train your advisor: Steering black-box {LLMs} with advisor models},
  author={Asawa, Parth and Zhu, Alan and O'Neill, Abby and Zaharia, Matei and Dimakis, Alexandros G and Gonzalez, Joseph E},
  journal={arXiv preprint arXiv:2510.02453},
  year={2025}
}

@article{zelikman2022star,
  title={{STaR}: Bootstrapping reasoning with reasoning},
  author={Zelikman, Eric and Wu, Yuhuai and Mu, Jesse and Goodman, Noah},
  journal={Advances in Neural Information Processing Systems},
  volume={35},
  pages={15476--15488},
  year={2022}
}

@article{yang2025qwen3,
  title={Qwen3 technical report},
  author={Yang, An and Li, Anfeng and Yang, Baosong and Zhang, Beichen and Hui, Binyuan and Zheng, Bo and Yu, Bowen and Gao, Chang and Huang, Chengen and Lv, Chenxu and others},
  journal={arXiv preprint arXiv:2505.09388},
  year={2025}
}

@article{singh2025openai,
  title={{OpenAI} {GPT-5} system card},
  author={Singh, Aaditya and Fry, Adam and Perelman, Adam and Tart, Adam and Ganesh, Adi and El-Kishky, Ahmed and McLaughlin, Aidan and Low, Aiden and Ostrow, AJ and Ananthram, Akhila and others},
  journal={arXiv preprint arXiv:2601.03267},
  year={2025}
}

@misc{google2026gemini31flashlite,
  title        = {Gemini 3.1 Flash-Lite Preview},
  author       = {{Google}},
  year         = {2026},
  howpublished = {\url{https://ai.google.dev/gemini-api/docs/models/gemini-3.1-flash-lite-preview}},
  note         = {Accessed: 2026-05-04}
}

@inproceedings{joshi-etal-2025-improving,
    title = "Improving Language Model Personas via Rationalization with Psychological Scaffolds",
    author = "Joshi, Brihi  and
      Ren, Xiang  and
      Swayamdipta, Swabha  and
      Koncel-Kedziorski, Rik  and
      Paek, Tim",
    booktitle = "Findings of the Association for Computational Linguistics: EMNLP 2025",
    year = "2025"
}

@article{hu2022lora,
  title={{LoRA}: Low-rank adaptation of large language models.},
  author={Hu, Edward J and Shen, Yelong and Wallis, Phillip and Allen-Zhu, Zeyuan and Li, Yuanzhi and Wang, Shean and Wang, Liang and Chen, Weizhu and others},
  journal={Iclr},
  volume={1},
  number={2},
  pages={3},
  year={2022}
}

@misc{oyez,
  author       = {{Oyez}},
  title        = {Oyez},
  year         = {2026},
  howpublished = {\url{https://www.oyez.org/}},
  note         = {Accessed: 2026-05-06}
}

@inproceedings{aher2023using,
  title={Using large language models to simulate multiple humans and replicate human subject studies},
  author={Aher, Gati V and Arriaga, Rosa I and Kalai, Adam Tauman},
  booktitle={International conference on machine learning},
  year={2023},
}

@inproceedings{park2023generative,
  title={Generative agents: Interactive simulacra of human behavior},
  author={Park, Joon Sung and O'Brien, Joseph and Cai, Carrie Jun and Morris, Meredith Ringel and Liang, Percy and Bernstein, Michael S},
  booktitle={Proceedings of the 36th annual acm symposium on user interface software and technology},
  year={2023}
}

@article{miroyan2025parastudent,
  title={ParaStudent: Generating and Evaluating Realistic Student Code by Teaching LLMs to Struggle},
  author={Miroyan, Mihran and Niousha, Rose and Gonzalez, Joseph E and Ranade, Gireeja and Norouzi, Narges},
  journal={arXiv preprint arXiv:2507.12674},
  year={2025}
}
\bibliographystyle{plainnat} 


\newpage
\appendix
\section{Ethics Statement}
\label{app:ethics_statement}

Our work uses real human interaction data for user modeling experiments. The data consists of (i) publicly available transcripts of public figures speaking in public forums and (ii) pseudonymous user-generated content from public online platforms. All data is already publicly accessible and does not include private communications or personally identifiable information beyond what is voluntarily disclosed in public settings. We do not attempt to de-anonymize users or infer sensitive attributes. Moreover, we do not plan on releasing our curated datasets or trained models. We replace the private Reddit usernames with User A/B in the paper. We acknowledge that pseudonymization and withholding artifacts reduce, but do not eliminate, re-identification and impersonation risks.

While user modeling and simulation raise broader ethical concerns, including potential misuse for impersonation or behavioral manipulation, our work focuses on controlled research settings aimed at improving modeling fidelity and explainability and does not deploy or evaluate systems in real-world decision-making contexts involving individuals.

\section{Data Processing}
\label{app:data}

This appendix describes the data sources, segmentation, and splits underlying the eight personas studied in Section~\ref{sec:experiments}, and reports per-persona dataset statistics.

\subsection{Format and Splits}
\label{app:data:format}

For each domain, raw source material is segmented into \emph{sessions}: one oral argument (SCOTUS), one PMQ sitting (PMQ), one episode (Podcasts), or one comment thread (Reddit). Each session is an ordered sequence of \texttt{(author, action)} turns. We then treat each turn produced by the target user as an action $a^*$ whose preceding turns within that session form the context $c$.

\paragraph{Session-level splits.}
Splits are assigned at the session level so that no context turn is shared between training and test data. For each persona, we shuffle eligible sessions with a fixed seed (42) and greedily allocate them in order to a training and test set, stopping each split once it reaches a target count of valid target-user turns. If a single session would exceed the remaining budget, only its prefix up to (and including) the last needed target turn is retained.

\paragraph{Turn filters.}
For SCOTUS, PMQ, and podcast transcripts, target turns whose action text matches the inaudible-marker pattern \texttt{"(?i)inaudible"} or contains the speaker-interruption marker \texttt{"——"} are excluded from both per-split budgeting and from generation and judging. Reddit comments are not subject to either filter.

\paragraph{Training corpus and GRPO subset.}
The training corpus is internally split into two disjoint subsets that we refer to as \texttt{train\_0} and \texttt{train\_1}. Their union is the retrieval pool from which all in-context demonstrations are drawn. \methodtrain{} and the \methodbaselinetrained{} baseline use only \texttt{train\_1} as their RL training set; \texttt{train\_0} is held out from RL training so that retrieval-time demonstrations include sessions the GRPO policy has not been optimized on. Test sessions are held out from both retrieval and training.

\subsection{Domain-Specific Sources}
\label{app:data:pipelines}

\paragraph{SCOTUS (Antonin Scalia, William Brennan).}
We use the Oyez Project's~\citep{oyez} oral argument transcripts via the public \texttt{walkerdb/supreme\_court\_transcripts} mirror (\url{https://github.com/walkerdb/supreme_court_transcripts}). Within each transcript, we extract speaker turns, drop empty turns, and merge consecutive turns by the same speaker. Justices are identified by Oyez's role annotation. To keep each Justice within a stylistically coherent period, we apply a date filter on the argued date: Antonin Scalia is restricted to 2007-01-01 through 2015-01-01, and William Brennan to 1969-01-01 through 1976-01-01. We do not plan to share the data publicly; it is collected for research purposes only.

\paragraph{PMQ (David Cameron, Tony Blair).}
PMQ sittings are scraped from the official UK Parliament Hansard API (\url{https://hansard-api.parliament.uk}). For each PM, we restrict to Commons sittings during their tenure (the half-open interval from their first day in office to their successor's first day) that contain a debate section titled ``Prime Minister''. We strip Hansard editorial annotations, drop chair (Speaker) interjections, normalize the PM's display name to a single canonical form, and discard procedural fragments (time stamps, ``rose---'' markers, and the boilerplate ``the Prime Minister was asked'' header). Sittings in which the PM did not personally answer (e.g., the Deputy PM substituting) are dropped. We do not plan to share the data publicly; it is collected for research purposes only.

\paragraph{Podcasts (Lex Fridman, Tim Ferriss).}
We use the official episode transcripts published by each host: \emph{Lex Fridman Podcast} transcripts from \url{https://lexfridman.com/podcast} and \emph{The Tim Ferriss Show} transcripts from \url{https://tim.blog/podcast}. Each episode forms one session. We normalize whitespace, drop empty turns, and merge consecutive turns by the same speaker. We do not plan to share the transcripts publicly; they are collected for research purposes only.

\paragraph{Reddit (User A, User B).}
We identify two of the most active users in the r/AmItheAsshole subreddit and download their full public comment history via the Arctic Shift archive (\url{https://arctic-shift.photon-reddit.com/download-tool}), filter to comments in r/AmItheAsshole, and reconstruct one conversational thread per qualifying comment by walking the parent chain up to the root post. Missing ancestor comments and root posts are batch-fetched from the Arctic Shift API. Threads are de-duplicated to their \emph{deepest} target-user turn, so that no record is a strict prefix of another. Threads are discarded if (i) the root post is deleted or removed, (ii) the title contains a moderator removal notice, (iii) the thread has fewer than two turns, (iv) any non-target turn above the user's comment is deleted, or (v) the average turn length is below 20 characters. We do not plan to share the data publicly; it is collected for research purposes only.

\paragraph{Copyright and data availability.} Hansard material contains Parliamentary information licensed under the Open Parliament Licence v3.0. We acknowledge Oyez and the transcript mirror, the official podcast transcript publishers, and Reddit contributors and the Arctic Shift archive as the respective sources. Podcast transcripts and Reddit content remain subject to applicable copyright and reuse restrictions. We do not plan to redistribute the collected transcripts, Reddit comments, curated datasets, or trained models.

\subsection{Dataset Statistics}
\label{app:data:stats}

Table~\ref{tab:data_stats} reports per-persona statistics. Sessions vary widely in length across domains, so the number of sessions per persona differs even when target-turn counts are matched: SCOTUS oral arguments produce dozens to hundreds of target turns per session, and PMQ sittings contribute roughly twenty per session, while Reddit threads typically contribute a single target turn each. Average target-turn length spans more than an order of magnitude, from short SCOTUS questions (15--18 words on average) to long-form Reddit replies (User B averages over 200 words per turn).

\begin{table}[ht]
\centering
\caption{Per-persona dataset statistics. \textbf{Sess.}\ counts sessions assigned to each split. \textbf{Tgt.}\ counts target-user turns after applying the per-domain filters in Section~\ref{app:data:format}. \textbf{Ctx.}\ counts non-target turns within the same sessions. \textbf{GRPO} is the number of target turns in the disjoint \texttt{train\_1} subset of the training corpus, used as the RL training set; the full training corpus, including this subset, is used as the retrieval pool at evaluation time. \textbf{Avg.\ words} is the mean word count per filtered target turn across all splits. \textbf{Date range} spans the full corpus.}
\label{tab:data_stats}
\setlength{\tabcolsep}{4pt}
\footnotesize
\begin{tabular}{llrrrrrrrrl}
\toprule
& & \multicolumn{3}{c}{\textbf{Train}} & \textbf{GRPO} & \multicolumn{3}{c}{\textbf{Test}} & & \\
\cmidrule(lr){3-5} \cmidrule(lr){7-9}
\textbf{Domain} & \textbf{User} & \textbf{Sess.} & \textbf{Tgt.} & \textbf{Ctx.} & \textbf{Tgt.} & \textbf{Sess.} & \textbf{Tgt.} & \textbf{Ctx.} & \textbf{Avg.\ words} & \textbf{Date range} \\
\midrule
\multirow{2}{*}{SCOTUS}
  & Antonin Scalia  &   357 & 4{,}096 & 73{,}358 & 3{,}072 &  55 & 512 & 11{,}417 &  17.5 & 2008--2014 \\
  & William Brennan &   419 & 4{,}096 & 72{,}591 & 3{,}072 &  52 & 512 &  9{,}883 &  15.7 & 1970--1975 \\
\midrule
\multirow{2}{*}{PMQ}
  & Tony Blair      &   166 & 4{,}096 &  4{,}656 & 3{,}072 &  21 & 512 &     540 & 106.3 & 1997--2007 \\
  & David Cameron   &   131 & 4{,}096 &  4{,}668 & 3{,}072 &  16 & 511 &     567 &  99.9 & 2010--2016 \\
\midrule
\multirow{2}{*}{Podcasts}
  & Lex Fridman     &    20 & 4{,}021 &  4{,}325 & 3{,}007 &   3 & 502 &     509 &  32.0 & 2023--2025 \\
  & Tim Ferriss     &    35 & 4{,}082 &  4{,}283 & 3{,}067 &   5 & 512 &     507 &  35.4 & 2016--2026 \\
\midrule
\multirow{2}{*}{Reddit}
  & User A         & 3{,}697 & 4{,}096 &  7{,}371 & 3{,}072 & 466 & 512 &     941 &  43.6 & 2019--2025 \\
  & User B         & 4{,}025 & 4{,}096 &  5{,}083 & 3{,}072 & 503 & 512 &     631 & 203.9 & 2022--2026 \\
\bottomrule
\end{tabular}
\end{table}

\section{Implementation Details}
\label{app:method_details}

For all prompts in all methods, the conversation context is truncated to a 4096 token budget.

\subsection{\methodbaseline{}}

Below we present the prompts used to generate candidate reasoning traces from $M_r$ for the \methodbaseline{} baseline. Contexts are formatted in XML format, with each turn as an element labeled with the author. To mitigate bias from seeing the user's name, turns uttered by the user are labeled with ``I'' rather than their name. The ``preamble\_block'' is a task-specific context that describes the task (e.g., ``You are a U.S. Supreme Court Justice. Your current task is to reconstruct your internal thought process while you were participating in an oral argument on the U.S. Supreme Court. The provided current context is an oral argument in a case. You are also given your next-turn spoken response to the current conversation context.''). The ``response\_format'' is task-specific context that describes the medium of the conversation (e.g., ``It is a transcribed spoken turn in an oral conversation.'' or ``It is a comment on a post on r/AmItheAsshole.'').

We sampled from the reasoning generator models with temperature $0.7$ for the Qwen and Llama models, and with the default reasoning effort for GPT-5-mini.

\begin{tcolorbox}[breakable,enhanced,title={\faComments~ Prompt for Reasoning Synthesis},
  listing only,
  listing options={
    language=none,
    basicstyle=\ttfamily\tiny,
    breaklines=true,
    linewidth=\linewidth,
    frame=single
  }]

\{preamble\_block\}
\medskip

\#\# Input
\medskip

<|Start Context|>

\{context\}

<|End of Context|>
\medskip

<|Start Ground Truth Response|>

\{ground\_truth\_action\}

<|End of Ground Truth Response|>
\medskip

\#\# Instructions

You are given a context and a response. Your task is to reconstruct the internal thought process that naturally leads from reading the context to producing the response.
\medskip

The thought process should:

- Be written in first person, present tense, as you actively read and process the context

- Surface your immediate reactions, interpretations, and situational understanding

- Reflect your beliefs, values, assumptions, and stance as they become relevant

- Organically arrive at a communicative intent (e.g., what to say, why, and how) without referencing or quoting the response itself

- Be consistent with your prior exchanges in the provided context. Your previous turns in the provided context (if any) have been labeled with <turn author="I">. Do not get confused by other participants referring to themselves in the first person; only turns labeled with <turn author="I"> are you.

- Never reference, quote, or allude to the ground truth response — the thought process must read as if the response has not been written yet

The thought process should feel like a genuine internal monologue - not a post-hoc analysis of the response, but the reasoning that precedes and motivates it.
\medskip

\#\# Output format
\medskip

- Only output the thought process without any additional text or commentary.
\end{tcolorbox}

\subsection{\methodprompt{}}

To obtain more diversity in the candidate reasoning traces, we sample reasoning traces using the \methodbaseline{} prompt from the Qwen and Llama models with temperature $1.0$. As GPT-5-mini does not allow temperature specification, we retain the default sampling settings. Below, we present the Action Reconstruction prompt given to the action model to reconstruct the action from the context and candidate reasoning trace. The action reconstructor model still used the default sampling temperature of $0.7$.

\begin{tcolorbox}[breakable,enhanced,title={\faComments~ Prompt for Action Reconstruction},
  listing only,
  listing options={
    language=none,
    basicstyle=\ttfamily\tiny,
    breaklines=true,
    linewidth=\linewidth,
    frame=single
  }]

\{preamble\_block\}
\medskip

\#\# Input
\medskip

<|Start Context|>

\{context\}

<|End of Context|>
\medskip

<|Start Thought Process|>

\{candidate\_reasoning\}

<|End of Thought Process|>
\medskip

\#\# Instructions
\medskip

- Generate the speaker's response to the provided context.

- The generated response should be consistent with the provided thought process.

- The response should be from the speaker's point of view.

- Your previous turns in the provided context (if any) have been labeled with <turn author="I">. Do not get confused by other participants referring to themselves in the first person; only turns labeled with <turn author="I"> are you.

- \{response\_format\}
\medskip

\#\# Output format
\medskip

- Only output the response without any additional text or commentary.
\end{tcolorbox}

The preamble block consists of task-specific information describing the domain (e.g., ``You are a U.S. Supreme Court Justice participating in an oral argument on the US Supreme Court. The provided current context is an oral argument in a case.'') and the response format string describes the medium of the conversation (e.g., ``It is a transcribed spoken turn in an oral conversation.'').

The judge prompt given to Gemini-3.1 Flash Lite is below. All reasoning candidates are presented to the judge at once to provide a more consistent evaluation across all candidates. We use the default recommended sampling parameters.

\begin{tcolorbox}[breakable,enhanced,title={\faComments~ Prompt for Action Alignment Scoring},
  listing only,
  listing options={
    language=none,
    basicstyle=\ttfamily\tiny,
    breaklines=true,
    linewidth=\linewidth,
    frame=single,
    keepspaces=true
  }]

Score how well each generated response aligns with the ground truth response along three interpretive dimensions: **style**, **intent**, and **values**.

\medskip

Each **alignment\_score** must reflect both (a) how well the generated response matches the ground truth on that dimension and (b) any need to downweight the score because of **extra or conflicting** content relative to the ground truth---unsupported additions, contradictions, or stylistic or pragmatic moves that diverge in ways that are not harmless paraphrases.

\medskip

Consider extra or conflicting content and features in each generated response relative to the ground truth when assigning each score. Verbose generated responses that include content and features not supported by the ground truth should receive **lower** scores on the relevant dimensions than more concise responses that better capture the core meaning and style of the ground truth.

\#\# Input
\medskip

<|Start Context|>

\{context\}

<|End of Context|>
\medskip

<|Start Ground Truth Response|>

\{ground\_truth\_response\}

<|End of Ground Truth Response|>
\medskip

<|Start Generated Responses|>

\{generated\_responses\}

<|End of Generated Responses|>

\medskip

\#\# Evaluation

\medskip

\#\#\# Dimension definitions

\{dimension\_information\}

\medskip

\#\# Instructions

- Analyze the provided context and the ground truth response.

- Analyze all \{k\} generated responses and compare each one independently to the ground truth response.

- For each generated response and each of the three dimensions (**style**, **intent**, **values**):

\begin{itemize}[leftmargin=15pt]
    \item Identify a few **key points** in the **ground truth response** (and context when needed) that are salient for that dimension.
    \item Assign an **alignment\_score** in [0, 1] for how well the generated response matches the ground truth on that dimension, already incorporating penalties for unsupported extras, contradictions, or harmful verbosity as they affect this dimension.
    \item In **alignment\_rationale**, briefly list the key ground-truth points you used, then explain the score with concrete comparisons to the generated response, including how extras, conflicts, or verbosity affected the score if applicable.
\end{itemize}
  
- Score each generated response independently. Multiple responses may receive similar or identical scores.

\medskip

\#\#\# Score anchors

| Score | Meaning |

|------|--------|

| 1.0 | Matches the ground truth on this dimension very closely, without meaningful unsupported extras |

| 0.7-0.9 | Strong match with minor gaps or very mild extra content |

| 0.4-0.6 | Partial match or noticeable extras or tensions |

| 0.1-0.3 | Weak match or clear conflicting or unsupported material on this dimension |

| 0.0 | Missing, contradicted, or heavily misaligned on this dimension |

\medskip

\#\# Output format (JSON)

Return a JSON object whose keys are the response indices ``"1"'' ... ``"\{k\}"''. Each key-value pair must have this shape:

\begin{verbatim}
{
  "1": {
    "style":  {
      "alignment_rationale": string,
      "alignment_score": float
    },
    "intent": {
      "alignment_rationale": string,
      "alignment_score": float
    },
    "values": {
      "alignment_rationale": string,
      "alignment_score": float
    }
  }
}
\end{verbatim}

\end{tcolorbox}

The dimensions are defined as follows:

\begin{description}[labelindent=1.5em, leftmargin=1.5em]
\item[(1) \textbf{Style}:]The surface-level linguistic properties of the utterance, entirely independent of meaning. Considers sentence length and syntactic complexity, formality and register, vocabulary and lexical choices, use of hedges, discourse markers, and fillers, and overall tone (assertive, tentative, emotive, dry).

\item[\textbf{(2) Intent}:]The specific propositional and pragmatic content of the utterance as a response to the given context. Considers the information conveyed, the stance taken on the immediate topic, the speech act performed (assertion, concession, challenge, deflection, etc.), and how it addresses what the prior turn raised.

\item[(3) \textbf{Values}:]The high-level, stable outlook of the speaker as revealed by the utterance. Considers the beliefs and assumptions about the world the speaker treats as given, the values and moral or social principles that shape their framing, and the ideological or philosophical commitments implicit in how they engage with the topic — characteristics that would remain consistent across different conversations and topics.
\end{description}

\subsection{\methodtrain{}}

We perform GRPO with a group size of $4$ and train a LoRA adapter with a rank of $128$. The prompt given to the policy model is the same one used in \methodprompt{}. To encourage diversity in the rollouts, we again set sampling temperature to $1.0$. The action reconstructor model still used the default sampling temperature of $0.7$.

Reward is calculated from the scoring prompt in \methodprompt{}, as well as an LM-judge call to the same judge model to penalize candidate reasoning traces that repeat the action verbatim; the prompt for the latter judge is presented below. Reward is calculated as $1\cdot(\text{average alignment score}) - 2 \cdot (\text{verbatim repeat score})$.

\begin{tcolorbox}[breakable,enhanced,title={\faComments~ Prompt for Non-duplication Judge},
  listing only,
  listing options={
    language=none,
    basicstyle=\ttfamily\tiny,
    breaklines=true,
    linewidth=\linewidth,
    frame=single
  }]
You are given a response and reasoning trace. You need to determine whether the response **explicitly** appears in the reasoning trace.
\medskip

\#\# Input

<|Start Reasoning Trace|>

{reasoning\_trace}

<|End of Reasoning Trace|>
\medskip

<|Start Response|>

{ground\_truth\_response}

<|End of Response|>
\medskip

\#\# Instructions

- Check if the response text **explicitly** appears in the reasoning trace.

- True means the response text is present verbatim in the reasoning trace. False means it is not present verbatim.

- Consider **minor** formatting differences and paraphrases as matches.
\medskip

\#\# Output format (JSON):

\begin{verbatim}
{
    "rationale": <brief explanation of your analysis>,
    "found": <true or false>
}
\end{verbatim}

\end{tcolorbox}

We train for 1 epoch over the full retrieval corpus with a learning rate $1\cdot 10^{-5}$ and cosine learning rate decay. We train on NVIDIA GPUs from the Hopper and Blackwell families. Training takes approximately 12 hours on a single GPU.

\subsection{\methodbaselinetrained{}}

We use the same training settings as \methodtrain{}. The reward function is also the same, except there is no duplication penalty. The prompt for the policy model is provided below.

\begin{tcolorbox}[breakable,enhanced,title={\faComments~ Prompt for \methodbaselinetrained{}},
  listing only,
  listing options={
    language=none,
    basicstyle=\ttfamily\tiny,
    breaklines=true,
    linewidth=\linewidth,
    frame=single
  }]

\{preamble\_block\}
\medskip

\#\# Input
\medskip

<|Start Context|>

\{context\}

<|End of Context|>
\medskip

\#\# Instructions

You are now responding to the conversation. Your previous utterances have been labeled with <turn author="I">. Do not get confused by other participants referring to themselves in the first person; only turns labeled with <turn author="I"> are you. First write private reasoning inside a single <thought>...</thought> block, then write what you wish to say (no XML tags) for this turn.
\end{tcolorbox}

\section{Evaluation Details}
\label{app:evaluation_details}

We provide further details for the evaluation process used to report win rates from Section \ref{subsec:evaluation}.

\subsection{Formalization}
Let $f_b$ refer to \methodbaseline{} and $f$ refer to the reasoning synthesis method we are testing. For each test context $c_T$ and ground-truth action $a_T^*$, we 
\begin{enumerate}
    \item Retrieve context-action pairs: $\{(c_i, a_i^*)\}_{i=1}^k$.
    \item Obtain \methodbaseline{} augmentations: $\{\hat{r}_i^b\}_{i=1}^k = f_b(c, a^*)$.
    \item Obtain \methodbaseline{}-based action: $\hat{a}_T^b = M_a(\{(c_i, \hat{r}_i^b, a^*_i)\}_{i=1}^k\}, c_T)$.
    \item Obtain augmentations from $f$: $\{\hat{r}_i^f\}_{i=1}^k = f(c, a^*)$.
    \item Obtain $f$-based action: $\hat{a}_T^f = M_a(\{(c_i, \hat{r}_i^f, a^*_i)\}_{i=1}^k\}, c_T)$.
    \item Compare $\hat{a}_T^b$ and $\hat{a}_T^f$ for similarity to $a_T^*$.
\end{enumerate}
Step 6 is performed using an LM pairwise judge, specifically Gemini-3.1 Flash Lite with the default recommended sampling parameters. We describe the prompt below.

\subsection{Action Generation}
Following Equation \ref{eq:pipeline}, we provide the action generation model with the augmented retrieved examples and the current test context. A total of 8 examples are retrieved per test context. We use Qwen3-4B-Embedding as the embedding model and use the context as keys and queries. The prompt used by the action generation model is given below; the test context $c_T$ truncated to a 4096 token budget. We sample from action models using their default sampling parameters.

\begin{tcolorbox}[breakable,enhanced,title={\faComments~ Prompt for Action Generation},
  listing only,
  listing options={
    language=none,
    basicstyle=\ttfamily\tiny,
    breaklines=true,
    linewidth=\linewidth,
    frame=single
  }]

\{preamble\_block\}
\medskip

\#\# Input
\medskip

<|Retrieved Examples Start|>

\{retrieved\_examples\}

<|Retrieved Examples End|>
\medskip

<|Current Context Start|>

\{test\_context\}

<|Current Context End|>
\medskip

\#\# Instructions

You are a participant in the conversation in the Current Context. It is now your turn to speak in the conversation.
Your previous utterances have been labeled with <turn author="I">. It is possible you have yet to speak in the conversation, in which case no turns in the Current Context are labeled with <turn author="I">.
Do not get confused by other participants referring to themselves in the first person; only turns labeled with <turn author="I"> are you. 
\medskip

You are given examples of your past thought processes and responses in similar past contexts.
Analyze each example triplet (context, thought process, response) and generate a thought process and response for the **current** conversation context.
\medskip

IMPORTANT RULES FOR THE THOUGHT PROCESS:

- Study how the thought processes in the examples reason over their contexts

- Follow the same thinking pattern in your generated thought process
\medskip

IMPORTANT RULES FOR THE RESPONSE:

- \{response\_format\}

- It does NOT need to cover all points from the thought process

- Match the same LENGTH and STYLE as the example responses

- Study how example responses are formed from their thought processes and follow the same pattern
\medskip

\#\# Output Format:

- Enclose the generated thought process in <|Thought Process Start|>...<|Thought Process End|>

- Enclose the generated response in <|Response Start|>...<|Response End|>
\end{tcolorbox}

We use the same preamble blocks and response formats as the action reconstructor in Section \ref{app:method_details}.

\subsection{Judgement}
Each call to our LM pairwise judge receives four inputs: the conversation context $c_T$ truncated to a 4096 token budget, the ground-truth action $a_T^*$, the generated response for model A, and the generated response for model B.

\begin{tcolorbox}[breakable,enhanced,title={\faComments~ Evaluation Prompt},
  listing only,
  listing options={
    language=none,
    basicstyle=\ttfamily\tiny,
    breaklines=true,
    linewidth=\linewidth,
    frame=single
  }]

You are a meticulous evaluator.
You will be given a context, a ground truth response and two generated responses (A and B) to the provided context.
Your job is to analyze and compare the ground truth and generated responses within the provided context along the specified dimensions.
\medskip

\#\# Input

<|Context|>

\{context\}

<|End Context|>
\medskip

<|Ground Truth Response|>

\{action\}

<|End Ground Truth Response|>
\medskip

<|Generated Response A|>

\{generated\_action\_a\}

<|End Generated Response A|>
\medskip

<|Generated Response B|>

\{generated\_action\_b\}

<|End Generated Response B|>
\medskip

\#\# Evaluation
\medskip

\#\#\# Dimension definitions

\{dimensions\_information\}
\medskip

\#\#\# Scoring

- Dimension scoring
\begin{itemize}[leftmargin=15pt]
    \item Analyze the ground truth response within the provided context and compare how well **A** and **B** capture the ground truth on that dimension.
    \item Provide your analysis and rationale for your final decision for each dimension.
    \item If the dimension cannot be meaningfully assessed for the given ground truth response and context, set the **winner** to "NA".
    \item If the winner is clear, set the **winner** to "A" or "B".
    \item If the winner is not clear but both responses are generally aligned with the ground truth along the specified dimension, set the **winner** to "tie".
    \item If both responses are totally misaligned with the ground truth along the specified dimension, set the **winner** to "tie (bad)".
\end{itemize}

- Overall scoring
\begin{itemize}[leftmargin=15pt]
    \item After assessing the generated responses on the specified dimensions, provide your final decision on which response is closer to the ground truth overall, taking into account the dimension-level analyses.
    \item Provide your analysis and rationale for your final decision.
    \item If the winner is clear, set the **winner** to "A" or "B".
    \item If the winner is not clear but both responses are generally aligned with the ground truth overall, set the **winner** to "tie".
    \item If both responses are totally misaligned with the ground truth overall, set the **winner** to "tie (bad)".
\end{itemize}
\medskip

\#\#\# Instructions and considerations

* For dimension-level scoring, evaluate each dimension independently: it is possible for A to win on one dimension and B to win on another.

* For overall scoring, consider the dimension-level analyses holistically along with any other relevant features and factors that may not be fully captured by the dimensions.

* For both dimension-level and overall scoring, use "tie" when neither response is clearly better compared to the ground truth response, and "NA" when the dimension cannot be meaningfully assessed for the given ground truth response and context.

* In your rationale, be specific and concrete in citing evidence from the ground truth and generated responses to support your analysis and decisions.

* For both dimension-level and overall scoring, consider extra or conflicting content and features in the generated responses relative to the ground truth, such as unsupported additions or contradictions, and factor these into your analysis and decisions.

* Verbose generated responses that include content and features that are not supported by the ground truth should be penalized compared to more concise responses that better capture the core meaning and style of the ground truth.

* Provide your analysis and rationale in a clear and organized manner, using specific examples from the ground truth and generated responses to illustrate your points.
\medskip

\#\# Output format (JSON)
Return exactly these keys:

\begin{verbatim}
{
    "style": {
        "rationale": string,
        "winner": "A" | "B" | "tie" | "NA"
    },
    "intent": {
        "rationale": string,
        "winner": "A" | "B" | "tie" | "NA"
    },
    "values": {
        "rationale": string,
        "winner": "A" | "B" | "tie" | "NA"
    },
    "overall": {
        "rationale": string,
        "winner": "A" | "B" | "tie"
    }
}
\end{verbatim}
\end{tcolorbox}

We use the same dimension definitions described in Appendix \ref{app:method_details}. For each dimension, our judge selects one of five outcomes: \textit{Model A wins}, \textit{Model B wins}, \textit{Tie} (both responses are generally aligned with the ground truth), \textit{Tie - Bad} (both responses are misaligned), or \textit{N/A} (the dimension doesn't apply to the response). Our LM judge is instructed to evaluate each dimension independently, such that Model A may win on style while Model B may win on values, and vice versa. The final overall verdict on which model did better depends on a holistic dimension-level analysis, and any ties (good or bad) are excluded from our final win-rate calculations.

\paragraph{Bias mitigation.} To mitigate position bias, we uniformly randomize our A and B model assignments during every turn. After the judge returns its verdict, we undo any random swaps so that our stored winners are consistent.

\section{Validations}

We present a synthetic probe of the hypothesis in Equation~\ref{eq:validity}, human validation of final responses, and additional evaluation controls.

\subsection{Assumption Validation}
\label{app:assumption_validation}

To validate our assumption that providing ground-truth reasoning traces in retrieved contexts leads to better downstream action prediction, we perform a small experiment on the Podcast domain. As we do not have access to ground truth reasoning from the real people in any of our domains, we synthesize new ground truth reasoning and action from each retrieved context and test context using GPT-5-mini. The task then becomes modeling the behavior of GPT-5-mini.

We then compare two methods using Qwen3-8B as the action generator. First, we allow Qwen3-8B to synthesize reasoning for each retrieved context (with the GPT-5-mini ground truth action) and prompt Qwen3-8B with the augmented retrievals to generate an action. Second, we use the ground-truth reasoning traces generated by GPT-5-mini as the synthesized reasoning and again prompt Qwen3-8B with the augmented retrievals to generate an action. Finally, we compare each method's generated actions against each other to assess their similarity to the ground-truth GPT-5-mini actions.

Across the 307 non-tied judgments, the latter method, which used ground-truth reasoning traces, had a win rate of 56.4\% compared to the former method ($p=0.015$). This validates our assumption that synthesizing reasoning traces close to the hidden ground-truth reasoning leads to better downstream action prediction. We note that this probe does not establish that our reconstruction-based method recovers ground truth human latent reasoning.

\subsection{Evaluation Validation}
\label{subsec:human_validation}
We extracted 100 random user turn samples from our four domains (Supreme Court, UK Parliament, Reddit, and Podcasts), with 25 turns per domain. Two human raters evaluated GPT-5-mini \methodprompt{} and \methodbaseline{} responses. For each turn, human raters were shown six turns of previous context, the ground truth action, and two candidate models' generated actions (A and B). Their orders were randomized per turn to prevent positional bias. The validation process had four options the raters could choose from: \textit{Model A wins}, \textit{Model B wins}, \textit{Tie} (both model responses were aligned), or \textit{Tie - Bad} (neither response was aligned), to indicate which response better captured the target user's style, intent, values, and overall mimicry. Any dimensions that the LM judge labeled as \textit{N/A} were excluded from the validation. Afterwards, we computed alignment on the ``overall'' dimension between the two human raters and between each human rater and the original LM judge (Gemini-3.1-Flash-Lite). In our alignment calculation, we collapsed the \textit{Tie} and \textit{Tie - Bad} ratings into a singular \textit{No Winner} category to reflect that both variants intended the same directional outcome. According to the human raters, \methodprompt{} achieves an average win rate of 60.7\% over \methodbaseline{} (ties excluded). Human--human agreement is 68.0\% (Cohen's $\kappa=0.440$). Human--LM agreement averages 78.0\% across the two raters, with a mean Cohen's $\kappa$ of 0.640. Both agreement statistics include ties merged into the \textit{No Winner} category.

\subsection{Disjoint-Judge Evaluation}
\label{subsec:disjoint_validation}
We re-ran the Qwen3-8B evaluation with a disjoint judge family (GPT-5.6 Terra), using the original judge only for selection. \methodprompt{} retains a modest pooled advantage: 52.1\% over \methodbaseline{}, versus 54.7\% under the original judge. Table~\ref{tab:disjoint_judge} reports results across all four domains; only SCOTUS individually shows a clear advantage over 50\%. Overall agreement between judges is 77.9\% when both choose a winner, and 46.5\% when ties are included.

\begin{table}[t]
\centering
\small
\caption{Disjoint-judge win rates for Qwen3-8B \methodprompt{} against \methodbaseline{} using GPT-5.6 Terra. Ties are excluded.}
\label{tab:disjoint_judge}
\begin{tabular}{lr}
\toprule
Domain & Win rate \\
\midrule
SCOTUS & 55.5\% \\
PMQ & 49.7\% \\
Podcasts & 51.6\% \\
Reddit & 51.7\% \\
\midrule
Pooled & 52.1\% \\
\bottomrule
\end{tabular}
\end{table}

\subsection{Additional Ablations}
\label{subsec:rebuttal_controls}

\paragraph{No-reasoning RAG.} We ran a RAG baseline without synthesized reasoning to isolate the contribution of reasoning augmentation from retrieval. On SCOTUS with Qwen3-8B, \methodprompt{} achieves a 62.7\% win rate over this baseline.

\paragraph{Candidate budget.} We varied the number of reasoning candidates with Qwen3-8B on SCOTUS. Win rates for $N=2,4,8$ are 56.8\%, 61.3\%, and 60.3\%, respectively, indicating an increase from $N=2$ to $4$ and saturation beyond $N=4$.

\section{Detailed Results}
\label{app:results_details}

In this Appendix we present tables showing detailed results, breaking down results by individuals.

\begin{table}[hp]
  \centering
  \caption{Figure \ref{fig:recon} results. Win rates against \methodbaseline{} and total number of non-tied test contexts, per user.}
  \label{tab:rr-comparisons}
  \resizebox{\textwidth}{!}{
      \begin{tabular}{
          l
          S[table-format=1.3] S[table-format=3.0]
          S[table-format=1.3] S[table-format=3.0]
          S[table-format=1.3] S[table-format=3.0]
        }
        \toprule
        & \multicolumn{2}{c}{Qwen \methodprompt{}}
        & \multicolumn{2}{c}{GPT \methodprompt{}}
        & \multicolumn{2}{c}{Qwen \methodbaselinetrained{}} \\
        \cmidrule(lr){2-3} \cmidrule(lr){4-5} \cmidrule(lr){6-7}
        User & {Win Rate} & {$n$ (non-tied)} & {Win Rate} & {$n$ (non-tied)} & {Win Rate} & {$n$ (non-tied)} \\
        \midrule
        antonin\_scalia & 0.596 & 213 & 0.553 & 217 & 0.186 & 242 \\
        william\_brennan& 0.627 & 244 & 0.580 & 226 & 0.287 & 282 \\
        david\_cameron  & 0.545 & 303 & 0.562 & 356 & 0.451 & 397 \\
        tony\_blair     & 0.517 & 300 & 0.522 & 345 & 0.516 & 397 \\
        lex\_fridman    & 0.476 & 271 & 0.505 & 220 & 0.439 & 337 \\
        tim\_ferriss    & 0.535 & 226 & 0.476 & 145 & 0.487 & 271 \\
        User A          & 0.518 & 253 & 0.531 & 228 & 0.285 & 365 \\
        User B          & 0.582 & 261 & 0.522 & 253 & 0.345 & 400 \\
        \bottomrule
      \end{tabular}
    }
\end{table}

\begin{table}[hp]
  \centering
  \caption{Figure \ref{fig:cross_model} results. Win rates against \methodbaseline{} reasoning and total number of non-tied test contexts, per user.}
  \label{tab:act-comparisons}
  \begin{tabular}{
      l
      S[table-format=1.3] S[table-format=3.0]
      S[table-format=1.3] S[table-format=3.0]
    }
    \toprule
    & \multicolumn{2}{c}{\shortstack{GPT Action Model\\Qwen \methodprompt{} Reasoning}}
    & \multicolumn{2}{c}{\shortstack{Qwen Action Model\\GPT \methodprompt{} Reasoning}} \\
    \cmidrule(lr){2-3} \cmidrule(lr){4-5}
    User & {Win Rate} & {$n$ (non-tied)} & {Win Rate} & {$n$ (non-tied)} \\
    \midrule
    william\_brennan& 0.570 & 235 & 0.516 & 275 \\
    antonin\_scalia & 0.480 & 229 & 0.562 & 242 \\
    david\_cameron  & 0.497 & 358 & 0.482 & 330 \\
    tony\_blair     & 0.478 & 364 & 0.472 & 320 \\
    lex\_fridman    & 0.514 & 222 & 0.564 & 312 \\
    tim\_ferriss    & 0.522 & 180 & 0.538 & 240 \\
    User A          & 0.548 & 208 & 0.544 & 250 \\
    User B          & 0.523 & 266 & 0.535 & 282 \\
    \bottomrule
  \end{tabular}
\end{table}

\begin{table}[hp]
  \centering
  \caption{Figure \ref{fig:model_ablations} results. Win rates against Backward Synthesis reasoning and total number of non-tied test contexts, per user.}
  \label{tab:learned-recon-paired}
  \begin{tabular}{
      l
      l
      S[table-format=1.3] S[table-format=3.0]
      S[table-format=1.3] S[table-format=3.0]
    }
    \toprule
    & & \multicolumn{2}{c}{\methodprompt{}} & \multicolumn{2}{c}{\methodtrain{}} \\
    \cmidrule(lr){3-4} \cmidrule(lr){5-6}
    User & Reasoning Model & {Win Rate} & {$n$ (non-tied)} & {Win Rate} & {$n$ (non-tied)} \\
    \midrule
    david\_cameron & Qwen-14B & 0.510 & 302 & 0.504 & 333 \\
    david\_cameron & Qwen-8B  & 0.562 & 356 & 0.481 & 297 \\
    david\_cameron & Llama-8B & 0.545 & 319 & 0.580 & 338 \\
    david\_cameron & Qwen-4B  & 0.591 & 328 & 0.696 & 339 \\
    \addlinespace
    tony\_blair    & Qwen-14B & 0.487 & 314 & 0.550 & 300 \\
    tony\_blair    & Qwen-8B  & 0.522 & 345 & 0.518 & 299 \\
    tony\_blair    & Llama-8B & 0.542 & 308 & 0.548 & 314 \\
    tony\_blair    & Qwen-4B  & 0.597 & 303 & 0.704 & 351 \\
    \bottomrule
  \end{tabular}
\end{table}



\end{document}